\documentclass{article}
\usepackage{spconf,amsmath,graphicx}

\usepackage{stmaryrd}
\usepackage{bm}
\usepackage{amssymb}
\usepackage{booktabs}
\usepackage{graphicx}
\usepackage{float}
\usepackage{graphicx}
\usepackage{stfloats}
\usepackage{amsmath}
\usepackage{epstopdf}
\usepackage{algorithm,algorithmic}
\usepackage{multirow}
\usepackage{bbm}
\usepackage{framed}
\usepackage[colorlinks,linkcolor=red,anchorcolor=green,citecolor=green]{hyperref}
\usepackage[numbers,sort&compress]{natbib}
\usepackage{amsthm}

\newcommand{\ie}{{\em i.e.}}
\newcommand{\eg}{{\em e.g.}}

\newtheorem{corollary}{Corollary}
\newtheorem{theorem}{Theorem}
\newtheorem{proposition}{Proposition}
\newtheorem{conjecture}{Conjecture}

\title{A Comparative Study for the Nuclear Norms Minimization Methods}
\name{Zhiyuan~Zha$^{1, 2}$, Bihan~Wen$^{2}$, Jiachao~Zhang$^{3}$, Jiantao~Zhou$^{4}$ and Ce~Zhu$^{1}$}
\address{$^1${School of Information and Communication Engineering,}\\
{University of Electronic Science and Technology of China, Chengdu, 611731, China.}\\
$^2${School of Electrical and Electronic Engineering, Nanyang Technological University, 639798, Singapore.}\\
$^3$ Kangni Mechanical and Electrical Institute, Nanjing Institute of Technology, Nanjing 211167, China.\\
$^4${Department of Computer and Information Science,  University of Macau, Macau 999078, China.}
}
\begin{document}
%
\maketitle
\begin{abstract}
The nuclear norm minimization (NNM) is commonly used to approximate the matrix rank by shrinking all singular values equally. However, the singular values have clear physical meanings in many practical problems,  and NNM may not be able to faithfully approximate the matrix rank. To alleviate the above-mentioned limitation of NNM, recent studies have suggested that the weighted nuclear norm minimization (WNNM) can achieve a better rank estimation than NNM, which heuristically set the weight being inverse to the singular values. However, it still lacks a rigorous explanation why WNNM is more effective than NMM in various applications. In this paper, we analyze NNM and WNNM from the perspective of group sparse representation (GSR). Concretely, an adaptive dictionary learning method is devised to connect the rank minimization and GSR models. Based on the proposed dictionary, we prove that NNM and WNNM are equivalent to $\ell_1$-norm minimization and the weighted $\ell_1$-norm minimization in GSR, respectively. Inspired by enhancing sparsity of the weighted $\ell_1$-norm minimization in comparison with $\ell_1$-norm minimization in sparse representation, we thus explain that WNNM is more effective than NMM. By integrating the image nonlocal self-similarity (NSS) prior with the WNNM model, we then apply it to solve the image denoising problem. Experimental results demonstrate that  WNNM is more effective than NNM and outperforms several state-of-the-art methods in both objective and perceptual quality.
\end{abstract}


\section{Introduction}
\label{sec:intro}

Due to the fact that the data from many practical cases has low-rank property, such as video \cite{1} and hyperspectral image \cite{2}, low-rank matrix approximation (LRMA) has shown great potentials in various applications including image processing \cite{4,5}, computer vision \cite{3,6,12} and machine learning \cite{7,8,9,10,11}. For instance, the foreground and background in a video are modeled as low-rank and sparse \cite{3,6}, respectively. The Netflix customer data matrix is treated as low-rank, since the customers' behaviors and attributes are highly correlated, thus can be modeled by a few principle components. 

Generally speaking, there are two popular approaches for LRMA: 
low-rank matrix factorization (LRMF) \cite{7,8} and rank minimization \cite{4,5,9,10,11,12}. 
In this paper, we focus on the rank minimization methods, with the nuclear norm minimization (NNM) \cite{9,11} being the representative one.
The goal of NNM is to recover the underlying low-rank matrix $\textbf{\emph{X}}$ from its degraded observation $\textbf{\emph{Y}}$, by minimizing the nuclear norm of $\textbf{\emph{X}}$. 
In recent years, various applications based on NNM have been developed, such as image/video denoising \cite{1,5}, background extraction \cite{3,6} and face shadow removal \cite{12.1,13}. However, the nuclear norm is usually adopted as a convex surrogate of the matrix rank. Though possessing the theoretical guarantee, NNM tries to shrink different rank components equally, and therefore it cannot estimate the matrix rank accurately enough. To mitigate the disadvantage of NNM, a variety of enhanced low-rank approximation methods have been proposed \cite{4,10,12,13,14,15}. In particular, the most well-known one is the weighted nuclear norm minimization (WNNM) model \cite{4,12}, which assigns different weights to different singular values such that the matrix rank approximation becomes more accurate than NNM. However, it  still lacks a rigorous explanation why WNNM is more effective than NNM in various applications.

Based on the above concern in mind, this paper explains why WNNM is more effective than NNM model from the point of group sparse representation (GSR).  To the best of our knowledge, this is the first work to propose a rigorous explanation why WNNM is more effective than NMM. Specifically, we firstly devise an adaptive dictionary learning method to connect the rank minimization and GSR models. Then, we prove that under the proposed dictionary, NNM and WNNM are equivalent to $\ell_1$-norm minimization and the weighted $\ell_1$-norm minimization in GSR, respectively. Encouraged by enhancing sparsity of the weighted $\ell_1$-norm minimization in comparison with $\ell_1$-norm minimization in sparse representation, we then indicate that WNNM is more effective  than NNM. 
We employ the WNNM model to image denoising along with image nonlocal self-similarity (NSS) \cite{25} prior. Experimental results demonstrate that WNNM is more effective than NNM and outperforms many state-of-the-art denoising methods.
\vspace{-2mm}
\section {Related Works}
\vspace{-2mm}
\subsection {Nuclear Norm Minimization}

According to \cite{9,11}, the nuclear norm is the tightest convex relaxation of the original rank minimization problem. Given a data matrix $\textbf{\emph{Y}}_i\in\mathbb{R}^{b\times c}$, NNM aims to find a matrix $\textbf{\emph{X}}_i\in\mathbb{R}^{b\times c}$ of rank $r$, which can be formulated as the following minimization problem,
\begin{equation}
\textstyle{\mathcal{D}_\lambda(\textbf{\emph{Y}}_i) =\arg\min_{\textbf{\emph{X}}_i}\frac{1}{2}\|\textbf{\emph{Y}}_i-\textbf{\emph{X}}_i\|_F^2 +\lambda\|\textbf{\emph{X}}_i\|_*},
\label{eq:1}
\end{equation} 
where $\|\textbf{\emph{X}}_i\|_*=\sum\nolimits_j{\sigma_{i, j}}$, and ${\sigma_{i, j}}$ is the $j$-th singular value of the matrix $\textbf{\emph{X}}_i$.  $\lambda$ is a positive constant.

\vspace{-2mm}
\subsection {Weighted Nuclear Norm Minimization}

Although a good theoretical guarantee by the singular value thresholding (SVT) model \cite{9}, NNM tends to over-shrink the rank components, and therefore, it achieves unsatisfactory accuracy for approximating the matrix rank. To improve the rank approximation performance of NNM,  Gu $\emph{et al}.$ \cite{4,12} proposed the WNNM model. To be concrete, the weighted nuclear norm $\|\textbf{\emph{X}}_i\|_{{\textbf{\emph{w}}_i},*}$  is used to regularize $\textbf{\emph{X}}_i$. Then, Eq.~\eqref{eq:1} can be rewritten as
\begin{equation}
\textstyle{\mathcal{D}_{{\textbf{\emph{w}}}_i} (\textbf{\emph{Y}}_i)=\arg\min_{\textbf{\emph{X}}_i}\frac{1}{2}\|\textbf{\emph{Y}}_i-\textbf{\emph{X}}_i\|_F^2 + \|\textbf{\emph{X}}_i\|_{{\textbf{\emph{w}}_i},*}}
\label{eq:3}
\end{equation} 
where $\|\textbf{\emph{X}}_i\|_{{\textbf{\emph{w}}_i},*}=\sum\nolimits_j{{{{w}}}_{i, j}}{\sigma_{i, j}}$,  ${\textbf{\emph{W}}}_i={\rm diag}({{{w}}}_{i, 1}, {{{w}}}_{i, 2},...,{{{w}}}_{i, j})$, $m ={\rm min (b, c)}$, $\forall j=1,\dots, m,$ and ${{{w}}}_{i, j}>0$ is a non-negative weight assigned to ${\sigma_{i, j}}$.

\vspace{-2mm}
\section {Analyzing NNM and WNNM based on the Group Sparse Representation}

In this section, we analyze NNM and WNNM from the point of GSR.  We first introduce the GSR model.

\vspace{-2mm}
\subsection {Group Sparse Representation}

\label{Sec:3.1}

Instead of using a single patch as the basic unit in patch sparse representation (PSR) \cite{16,17}, the GSR model considers each similar \emph{patch group} as the basic unit and has shown great potentials in various image processing tasks \cite{18,19,20,21,22}.  The GSR provides a powerful mechanism to integrate local sparsity and NSS of images. To be concrete, an image $\textbf{\emph{x}}$ with size $\sqrt{N}\times\sqrt{N}$ is divided into $n$ overlapped patches $\textbf{\emph{x}}_i$ of size $\sqrt{b} \times \sqrt{b}$, $i=1, 2, ..., n$. Then, we use each patch $\textbf{\emph{x}}_i$ as a reference patch, and impose an $L \times L$ sized searching window that is centered at  $\textbf{\emph{x}}_i$. K-Nearest Neighbour (KNN) algorithm \cite{23} is used to select $c$ most similar patches (based on  their Euclidean distance to the reference patch) to form a set $\textbf{\emph{S}}_i$. Following this, all the patches in $\textbf{\emph{S}}_i$ are stacked into a data matrix $\textbf{\emph{X}}_i \in{\mathbb R}^{b\times c}$, which contains each element of $\textbf{\emph{S}}_i$ as its column, \ie, ${\textbf{\emph{X}}}_i=\{{\textbf{\emph{x}}}_{i,1}, {\textbf{\emph{x}}}_{i,2},...,{\textbf{\emph{x}}}_{i, c}\}$. This matrix $\textbf{\emph{X}}_i$ consisting of patches with similar structures is thereby called a \emph{patch group},  where $\{{\textbf{\emph{x}}}_{i, j}\}_{j=1}^c$ denotes the $j$-th patch in the $i$-th \emph{patch group}.  Similar to PSR \cite{16,17}, given a dictionary ${\textbf{\emph{D}}}_{i}$, each  \emph{patch group} ${\textbf{\emph{X}}}_i$ can be sparsely represented by solving the following $\ell_0$-norm minimization problem,
\begin{equation}
\textstyle{\hat{\textbf{\emph{A}}}_i=\arg\min_{{\textbf{\emph{A}}}_i} \left(\frac{1}{2}\left\|{\textbf{\emph{X}}}_i-{\textbf{\emph{D}}}_i{\textbf{\emph{A}}}_i\right\|_F^2
+\lambda\left\|{\textbf{\emph{A}}}_i\right\|_0\right),}
\label{eq:5}
\end{equation} 
where ${\textbf{\emph{A}}}_i$ represents the group sparse coefficient of each \emph{patch group} ${\textbf{\emph{X}}}_i$. $\left\|~\right\|_F^2$ denotes the Frobenius norm, and $\left\|~\right\|_0$ signifies the $\ell_0$-norm, \ie, counting the nonzero entries of each column in ${\textbf{\emph{A}}}_i$.

However, since $\ell_0$-norm minimization problem is a difficult combinatorial optimization problem, solving Eq.~\eqref{eq:5} is NP-hard. Therefore, Eq.~\eqref{eq:5} is usually relaxed to the convex $\ell_1$-norm minimization counterpart, \ie,
\begin{equation}
\textstyle{\hat{\textbf{\emph{A}}}_i=\arg\min_{{\textbf{\emph{A}}}_i} \left(\frac{1}{2}\left\|{\textbf{\emph{X}}}_i-{\textbf{\emph{D}}}_i{\textbf{\emph{A}}}_i\right\|_F^2
+\lambda\left\|{\textbf{\emph{A}}}_i\right\|_1\right),}
\label{eq:6}
\end{equation} 

However, in some practical problems, such as image inverse problems \cite{1,20}, $\ell_1$-norm minimization is quite hard to achieve a sparse solution accurately, and thus leads to a poor reconstruction performance. This raises the question of whether we can improve the sparsity of  $\ell_1$-norm minimization or not. In other words, we wish that  $\ell_1$-norm minimization can be an alternative to  $\ell_0$-norm minimization and achieve a better solution. For this reason, Cand{\`e}s $\emph{et al}.$ \cite{24} proposed a well-known norm minimization method, \ie, the weighted $\ell_1$-norm minimization, and instead of Eq.~\eqref{eq:6}, we have the following minimization problem,
\begin{equation}
\textstyle{\hat{\textbf{\emph{A}}}_i=\arg\min_{{\textbf{\emph{A}}}_i} \left(\frac{1}{2}\left\|{\textbf{\emph{X}}}_i-{\textbf{\emph{D}}}_i{\textbf{\emph{A}}}_i\right\|_F^2
+\left\|{\textbf{\emph{W}}}_i\circ{\textbf{\emph{A}}}_i\right\|_1\right),}
\label{eq:7}
\end{equation} 
where $\circ$ represents the element-wise product of two matrices,  here ${\textbf{\emph{W}}}_i$ is a weight assigned to each ${\textbf{\emph{A}}}_i$ and it can enhance the representation capability of ${\textbf{\emph{A}}}_i$. Note that the weight ${\textbf{\emph{W}}}_i$ is inversely proportional to ${\textbf{\emph{A}}}_i$ \cite{24}. Meanwhile, we have the following conjecture.

\begin{conjecture}
\label{proposition:1}
\cite{24} Defining the weighted $\ell_1$-norm minimization ${ \textbf{v}_1} =\arg\min_{{\textbf{{x}}}\in\mathbb{R}^{n}}\left\|{\textbf{{W}}}{\textbf{{x}}}\right\|_1$ and  $\ell_1$-norm minimization  ${ \textbf{v}_2} =\arg\min_{{\textbf{{x}}}\in\mathbb{R}^{n}}\left\|{\textbf{{x}}}\right\|_1$, then we have,
\begin{equation}
 \textstyle{{ \textbf{v}_1}\succ { \textbf{v}_2},}
\label{eq:8}
\end{equation} 
where the weight ${\textbf{{W}}}$ is inversely proportional to  $|{\textbf{{x}}}|$. ${ \textbf{v}_1}\succ { \textbf{v}_2}$ denotes that the entry ${ \textbf{v}_1}$ has  much more sparsity encouraging than the entry ${ \textbf{v}_2}$ .
\end{conjecture}


In order to explain why WNNM is more effective than NNM model, we analyze them from the point of GSR. We first introduce an adaptive dictionary learning method.
\vspace{-2mm}
\subsection {Adaptive Dictionary Learning}

In this subsection, we present an adaptive dictionary learning method. For each \emph{patch group} ${\textbf{\emph{X}}}_i$, its adaptive dictionary can be learned from its observation ${\textbf{\emph{Y}}}_i\in\mathbb{R}^{b \times c}$.
Specifically, we apply the singular value decomposition (SVD) to ${\textbf{\emph{Y}}}_i$,
\begin{equation}
\textstyle{{\textbf{\emph{Y}}}_i= {\textbf{\emph{U}}}_i{\boldsymbol\Delta}_i{\textbf{\emph{V}}}_i^T=\sum\nolimits_{j=1}^{m} \delta_{i, j}{\textbf{\emph{u}}}_{i, j}{\textbf{\emph{v}}}_{i, j}^T,}
\label{eq:9}
\end{equation}
where ${\boldsymbol\Delta}_i={\rm diag}(\delta_{i, 1}, \delta_{i, 2},..., \delta_{i, {m}})$ is a diagonal matrix, ${m}={\rm min}(b, c)$, $\forall j=1,\dots, m,$  and  ${\textbf{\emph{u}}}_{i, j}, {\textbf{\emph{v}}}_{i, j}$ are the columns of ${\textbf{\emph{U}}}_i$ and ${\textbf{\emph{V}}}_i$, respectively.

Following this, we define each dictionary atom $\textbf{\emph{d}}_{i, j}$ of the adaptive dictionary $\textbf{\emph{D}}_i$ for each \emph{patch group} $\textbf{\emph{Y}}_i$, namely, $\textbf{\emph{d}}_{i, j}={\textbf{\emph{u}}}_{i, j}{\textbf{\emph{v}}}_{i, j}^T$,  $\forall j=1,\dots, m$.
We then have learned an adaptive dictionary, \ie,
\begin{equation}
\textstyle{\textbf{\emph{D}}_i=[\textbf{\emph{d}}_{i, 1}, \textbf{\emph{d}}_{i, 2}, ... , \textbf{\emph{d}}_{i, {m}}]}
\label{eq:11}
\end{equation}

It can be seen that the proposed dictionary learning method only needs one SVD operation per \emph{patch group}.
\vspace{-2mm}
\subsection {WNNM is More Effective than NNM}

Now, recalling the adaptive dictionary defined in Eq.~\eqref{eq:11}, given the degraded matrix ${\textbf{\emph{Y}}}_i$, $\ell_1$-norm minimization based on GSR model can be represented as
\begin{equation}
\textstyle{\hat{\textbf{\emph{A}}}_i=\arg\min_{{\textbf{\emph{A}}}_i} \left(\frac{1}{2}\left\|{\textbf{\emph{Y}}}_i-{\textbf{\emph{D}}}_i{\textbf{\emph{A}}}_i\right\|_F^2
+\lambda\left\|{\textbf{\emph{A}}}_i\right\|_1\right),}
\label{eq:12}
\end{equation} 

Similarly, given the degraded matrix ${\textbf{\emph{Y}}}_i$, then the weighted $\ell_1$-norm minimization based on GSR model can be formulated as the following minimization problem,
\begin{equation}
\textstyle{\hat{\textbf{\emph{A}}}_i=\arg\min_{{\textbf{\emph{A}}}_i} \left(\frac{1}{2}\left\|{\textbf{\emph{Y}}}_i-{\textbf{\emph{D}}}_i{\textbf{\emph{A}}}_i\right\|_F^2+
\left\|{\textbf{\emph{W}}}_i\circ{\textbf{\emph{A}}}_i\right\|_1\right),}
\label{eq:13}
\end{equation} 

According to the above design of the adaptive dictionary ${\textbf{\emph{D}}}_i$ in Eq.~\eqref{eq:11}, we have the following conclusions.

\begin{theorem}
\label{theorem:4}
The NNM in Eq.~\eqref{eq:1} is equivalent to the $\ell_1$-norm minimization in Eq.~\eqref{eq:12} under the proposed adaptive dictionary ${\textbf{{D}}}_i$.
\end{theorem}
\begin{corollary}
\label{corollary:1}
The WNNM in Eq.~\eqref{eq:3} is equivalent to the weighted $\ell_1$-norm minimization in Eq.~\eqref{eq:13} under the proposed adaptive dictionary ${\textbf{{D}}}_i$.
\end{corollary}

Then, based on Theorem~\ref{theorem:4}, Corollary~\ref{corollary:1},  and assuming the correctness of Conjecture~\ref{proposition:1}, we have the following proposition.
\begin{proposition}
\label{proposition:2}
Defining the weighted nuclear norm minimization $\textbf{h}_1 = \arg\min_{\textbf{{X}}_i} \left\|\textbf{{X}}_i\right\|_{\rm {\textbf{w}_i},*}$ and the nuclear norm minimization $\textbf{h}_2 = \arg\min_{\textbf{{X}}_i} \left\|\textbf{{X}}_i\right\|_{*}$, then we have
\begin{equation}
\textstyle{\textbf{h}_1 \gg  \textbf{h}_2.}
\label{eq:15}
\end{equation} 
where ${ \textbf{h}_1}\gg { \textbf{h}_2}$ denotes that the entry ${ \textbf{h}_1}$ is more effective than the corresponding entry ${ \textbf{h}_2}$. 
It is noted that the weight is inversely proportional to the singular value of $\textbf{{X}}_i$.
\end{proposition}

Based on Proposition~\ref{proposition:2}, we have thus explained why WNNM is more effective than NNM model. Note that, there are a variety of methods to devise the dictionary and the proposed adaptive dictionary learning approach is just one of them. Though the proposed dictionary learning seems to directly translate the sparse representation into the rank minimization problem, the main difference between the sparse representation and rank minimization models (\eg, NNM \cite{9} and WNNM \cite{4}) is that sparse representation has a dictionary learning operator, while the rank minimization problem does not, to the best of our knowledge.

\vspace{-2mm}
\section{GSR-WNNM for Image Denoising}

Since the GSR model is exploited to analyze the effectiveness of WNNM, we termed the proposed scheme as GSR-WNNM. In this section, to validate the effectiveness of WNNM model, we employ the GSR-WNNM model to solve the image denoising problem. Mathematically, image denoising \cite{26,27,28,29,41} aims to restore the clean  image $\textbf{\emph{x}}$ from its noisy observation $\textbf{\emph{y}}$, which can be generally modeled as $\textbf{\emph{y}}= \textbf{\emph{x}} + \textbf{\emph{n}}$, where $\textbf{\emph{n}}$ is usually assumed to be additive white Gaussian noise. 
In this work, we use the image NSS prior \cite{25} with the GSR-WNNM model to image denoising.  
The NSS prior indicates that many similar patches can be searched for any exemplar patch by natural images within nonlocal regions.
Specifically, each patch $\textbf{\emph{y}}_i$ is extracted from the degraded image $\textbf{\emph{y}}$. Like in the subsection~\ref{Sec:3.1}, we search for its $c$ nonlocal similar patches to generate a \emph{patch group} $\textbf{\emph{Y}}_i$, \ie, ${\textbf{\emph{Y}}}_i = \{{\textbf{\emph{y}}}_{i, 1}, {\textbf{\emph{y}}}_{i, 2},...,{\textbf{\emph{y}}}_{i, c}\}$. Then we have ${\textbf{\emph{Y}}}_i = \textbf{\emph{X}}_i +{\textbf{\emph{N}}}_i$, where $\textbf{\emph{X}}_i$ and ${\textbf{\emph{N}}}_i$ are the group matrices of original image and noise, respectively. Since all the patches in each data matrix have similar structures, the constructed data matrix ${\textbf{\emph{Y}}}_i$ has a low-rank property and we can use the low-rank approximation method to estimate $\textbf{\emph{X}}_i$ from $\textbf{\emph{Y}}_i$.  Then, by invoking the GSR-WNNM model, each \emph{patch group} $\textbf{\emph{X}}_i$ can be reconstructed by solving the following optimization problem,
\begin{equation}
\textstyle{\hat{\textbf{\emph{X}}}_i=\arg\min_{\textbf{\emph{X}}_i}\left(\frac{1}{2}\left\|\textbf{\emph{Y}}_i-\textbf{\emph{X}}_i\right\|_F^2 + \left\|\textbf{\emph{X}}_i\right\|_{ {\emph{\textbf{w}}}_i,*}\right).}
\label{eq:16}
\end{equation} 

Then, the closed-form solution of Eq.~\eqref{eq:16} can be solved by
\begin{equation}
\textstyle{\sigma_{i, j}= {\rm soft} (\delta_{i, j}, w_{i, j}) = {\rm max} (\delta_{i, j} - w_{i, j}, 0),}
\label{eq:17}
\end{equation}
where $\textbf{\emph{Y}}_i= \textbf{\emph{U}}_i\boldsymbol\Delta_i\textbf{\emph{V}}_i^T$ is the SVD of $\textbf{\emph{Y}}_i\in{\mathbb R}^{b\times c}$ with $\boldsymbol\Delta_i= {\rm diag}(\delta_{i, 1},...,\delta_{i, m})$, $ m = {\rm min}(b, c)$, $\forall j=1,\dots, m.$ $\textbf{\emph{X}}_i = \textbf{\emph{U}}_i\boldsymbol\Sigma_i\textbf{\emph{V}}_i^T$ is the SVD of $\textbf{\emph{X}}_i$ with $\boldsymbol\Sigma_i = {\rm diag}(\sigma_{i, 1},...,\sigma_{i, m})$. 

With the solution of $\boldsymbol\Sigma_i$ in Eq.~\eqref{eq:17}, the clean group matrix $\hat{{\textbf{\emph{X}}}}_i$ can be recovered as ${{\hat{\textbf{\emph{X}}}}_i}={{\textbf{\emph{U}}}_i}\boldsymbol\Sigma_i{{\textbf{\emph{V}}}_i^T}$. Then the clean image ${{\hat{\textbf{\emph{x}}}}}$ can be reconstructed by aggregating all the groups $\{{\hat{\textbf{\emph{X}}}_i}\}$.

The weight $\textbf{\emph{W}}_i$ of each \emph{patch group} $\textbf{\emph{X}}_i$ is usually set to be inversed to the singular values, and thus, in \cite{4}, the weight is heuristically set as $w_{i, j} = q/(\sigma_{i, j}+\varepsilon)$, where $q$ and $\varepsilon$ are the constant. However, WNNM model in \cite{4} sometimes ejects the terminating error. In this paper, we present an adaptive weight setting scheme to avoid this error. Specifically, inspired by \cite{30}, the weight $\textbf{\emph{W}}_i$ of each \emph{patch group} $\hat{\textbf{\emph{X}}}_i$ is set as $\textbf{\emph{w}}_i = [w_{i,1}, w_{i,2}, ..., w_{i, j}]$. We have $\textbf{\emph{w}}_i={q* 2\sqrt{2}{\sigma_n}^2}/{(\boldsymbol\gamma_i+\varepsilon)}$, where $\sigma_n^2$ represents the noise variance, and $\boldsymbol\gamma_i$ denotes the estimated standard variance of the singular values of each \emph{patch group}. The complete description of the GSR-WNNM for image denoising is exhibited in Algorithm~\ref{algo:1}.

\begin{algorithm}[htbp]
\small
\caption{The GSR-WNNM for image denoising.}
\begin{algorithmic}[1]
\REQUIRE Noisy image $\textbf{\emph{y}}$.
\STATE  Initialize $\hat{\textbf{\emph{x}}}^{0}=\textbf{\emph{y}}$, $\textbf{\emph{y}}^{0}=\textbf{\emph{y}}$. 
\FOR{$k=0$ \TO $K$}
\STATE  Iterative regularization $\textbf{\emph{y}}^{k}=\hat{\textbf{\emph{x}}}^{k-1}+ \mu (\textbf{\emph{y}}-\hat{\textbf{\emph{x}}}^{k-1})$;
\FOR{Each patch ${\textbf{\emph{y}}}_{i}$ in ${\textbf{\emph{y}}}^k$ }	
\STATE Find nonlocal similar patches to form a \emph{patch group} ${\textbf{\emph{Y}}}_{i}$;
\STATE Singular value decomposition $[\textbf{\emph{U}}_i, \boldsymbol\Delta_i, \textbf{\emph{V}}_i]= SVD ({\textbf{\emph{Y}}}_i)$;
\STATE Estimate the weight $\textbf{\emph{w}}_i$ of each \emph{patch group} by compute   $\textbf{\emph{w}}_i={q* 2\sqrt{2}{\sigma_n}^2}/{(\boldsymbol\gamma_i+\varepsilon)}$;
\STATE Calculate $\boldsymbol\Sigma_i$ by  Eq.~\eqref{eq:17};
\STATE Reconstruct   $\hat{\textbf{\emph{X}}}_i=\textbf{\emph{U}}_i\boldsymbol\Sigma_i\textbf{\emph{V}}_i^T$;
\ENDFOR
\STATE Aggregate ${\textbf{\emph{X}}}_i$  to form the restored image $\hat{\textbf{\emph{x}}}^{k}$.
\ENDFOR
\STATE $\textbf{Output:}$ The final denoised image $\hat{\textbf{\emph{x}}}$.
\end{algorithmic}
\label{algo:1}
\end{algorithm}

\begin{figure}[!htbp]
	\vspace{-4mm}
\begin{minipage}[b]{1\linewidth}
\centering
\centerline{\includegraphics[width=8.5cm]{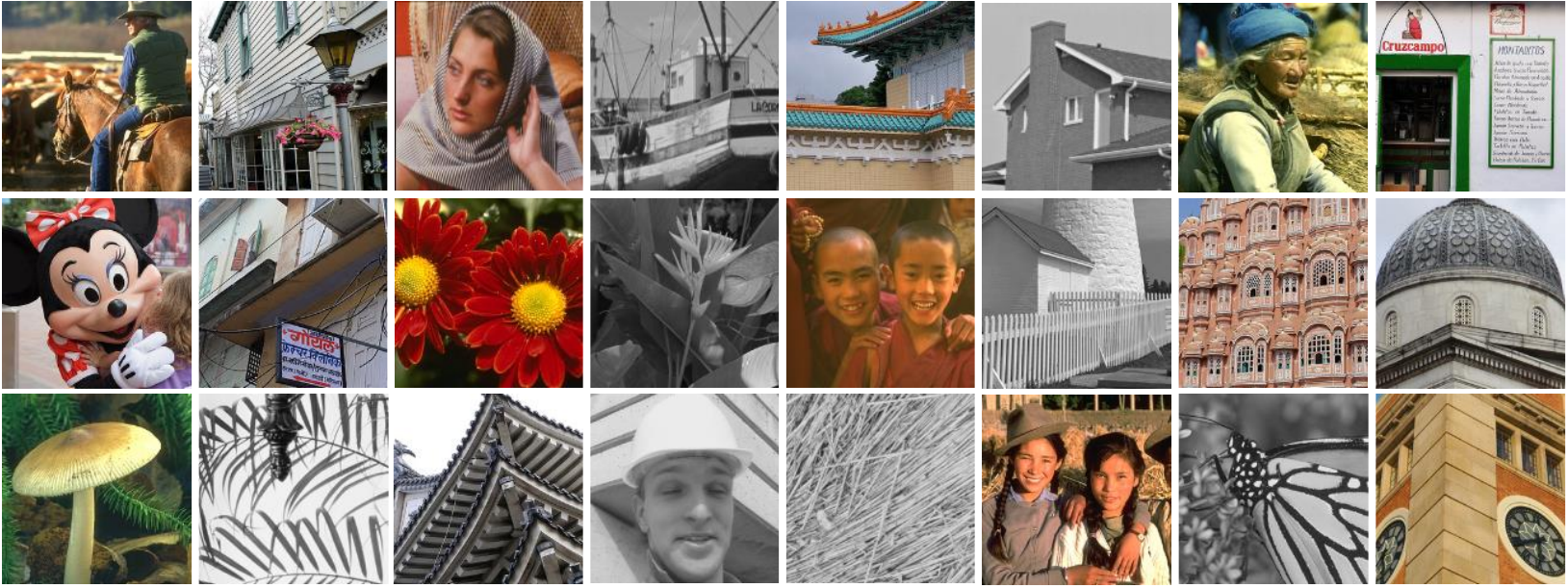}}
\end{minipage}
\vspace{-6mm}
\caption{ All test images.}
\label{fig:1}
\vspace{-4mm}
\end{figure}

\vspace{-2mm}
\section{Experimental Results}

In this section, we conduct experiments to validate the performance of the GSR-WNNM for image denoising. To verify the effectiveness of WNNM model, we compare it with NNM model (dubbed GSR-NNM in our work). All experimental test images are shown in Fig.~\ref{fig:1}. The source code is available at: \url{https://drive.google.com/open?id=1ZAkBCMPxIzgZ36y4zs-kg9ot7kqXnuHD}.

\begin{figure}[!htbp]
\vspace{-4mm}
	\centerline{\includegraphics[width=8.5cm]{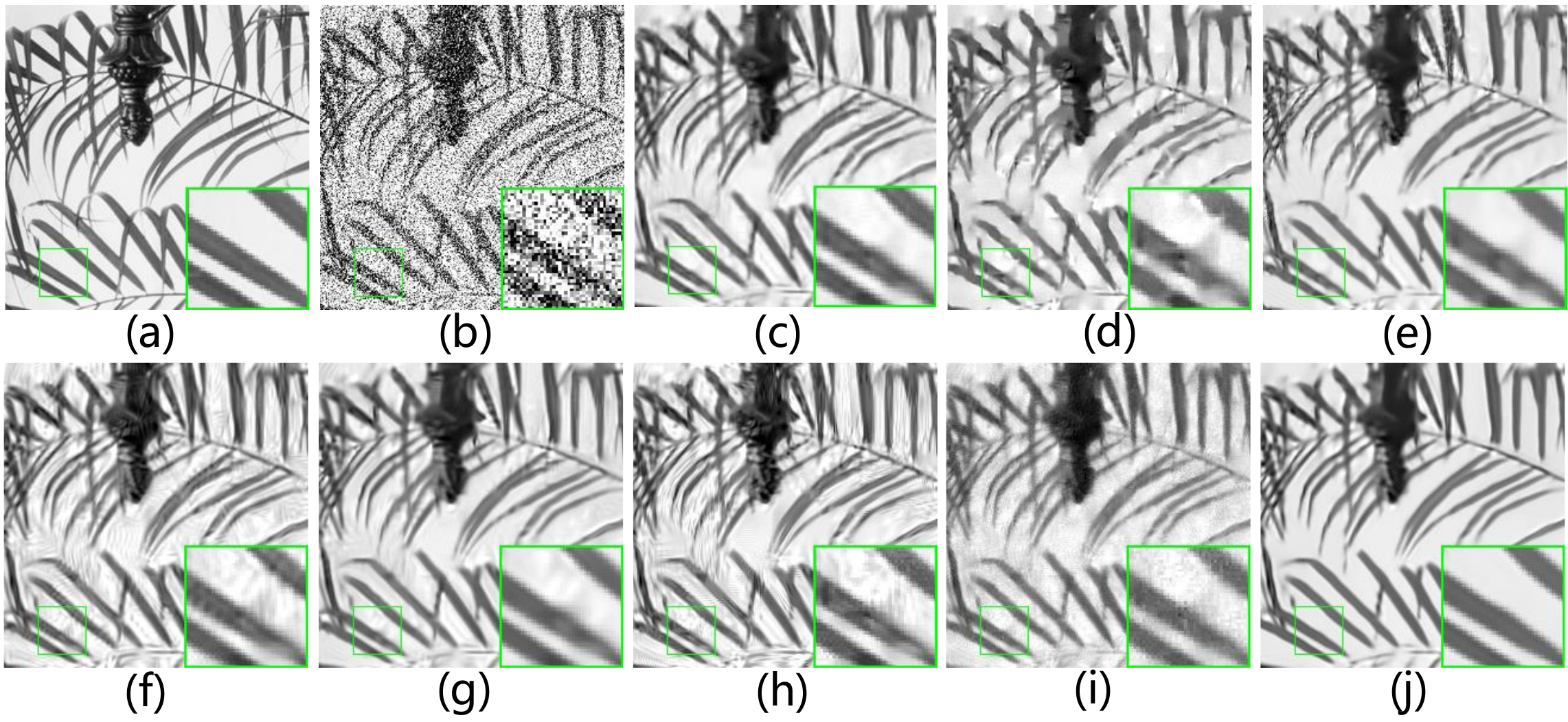}}
	\vspace{-3mm}
	\caption{\footnotesize  Denoising performance comparison of image $\emph{Leaves}$ with $\sigma_n$ = 100. (a) Original image; (b) Noisy image; (c) BM3D (PSNR = 20.90dB); (d) EPLL  (PSNR = 20.26dB); (e) NCSR (PSNR = 20.84dB); (f) Plow  (PSNR = 20.43dB); (g) PGPD  (PSNR = 20.95dB);  (h) OGLR (PSNR = 20.28dB);   (i) GSR-NNM (PSNR = 19.57dB); (j) GSR-WNNM (PSNR = \textbf{21.56dB}).}
\label{fig:2}
\vspace{-4mm}
\end{figure}

We compare the GSR-WNNM with some leading denoising methods, including BM3D \cite{18}, EPLL \cite{26},  Plow \cite{28}, NCSR \cite{27}, PGPD \cite{21}, OGLR \cite{29} and GSR-NNM methods. The averaged PSNR values of the GSR-WNNM, as well as the selected competing methods are shown in Table~\ref{tab:1}. The averaged PSNR gains of the GSR-WNNM over BM3D, EPLL, NCSR, Plow, PGPD, OGLR and GSR-NNM methods are 0.32dB, 0.82dB, 0.44dB, 0.85dB, 0.21dB, 0.55dB and 1.50dB, respectively. It is clear that the GSR-WNNM significantly outperforms the GSR-NNM method. Therefore, this result is consistent with our above theoretical analysis.  A denoising example in the case of $\sigma_n$ =100 for image $\emph{Leaves}$ is shown in Fig.~\ref{fig:2}. It can be seen that  over-smooth effects or undesirable artifacts are generated by BM3D, EPLL, NCSR, Plow, PGPD, OGLR and GSR-NNM methods. The  GSR-WNNM improves the quality of the denoised images by significantly reducing artifacts and over-smooth effects.  
Therefore, these results demonstrate that WNNM is more effective than NNM model and also validate the superior of the WNNM model.

\begin{table}[!htbp]
\footnotesize
\vspace{-4mm}
\caption{Average PSNR results (dB) of different methods.}
\resizebox{0.48\textwidth}{!}
{
\centering  
\begin{tabular}{|c|c|c|c|c|c|c|c|}
\hline
\multirow{1}{*}{\textbf{{Noise Level}}}& \textbf{20}  &\textbf{30}  &\textbf{40}  &\textbf{50}   &\textbf{75} &\textbf{100}\\
 \hline
  \multirow{1}{*}{BM3D \cite{18}} & 31.64 & 29.65 & 28.04 & 27.09 & 25.17 & 23.81  \\
\hline
  \multirow{1}{*}{EPLL \cite{26}} & 31.19 & 29.14 & 27.69 & 26.58 & 24.58 & 23.21 \\
\hline
 \multirow{1}{*}{Plow \cite{28}}  & 31.03 & 29.19 & 27.79 & 26.67 & 24.53 & 23.01  \\
\hline
 \multirow{1}{*}{NCSR \cite{27}}  & 31.65 & 29.54 & 28.10 & 26.99 & 24.92 & 23.45 \\
 \hline
  \multirow{1}{*}{PGPD \cite{21}} & 31.71 & 29.69 & 28.29 & 27.22 & 25.26 & 23.87 \\
\hline
 \multirow{1}{*}{OGLR \cite{29}}  & 31.47 & 29.54 & 28.14 & 26.85 & 24.87 & 23.14 \\
 \hline
 \multirow{1}{*}{GSR-NNM}  & 30.29 & 28.37 & 27.52 & 26.14 & 23.92 & 22.04 \\
 \hline
 \multirow{1}{*}{GSR-WNNM} & \textbf{31.89} & \textbf{29.91} & \textbf{28.51} & \textbf{27.41} & \textbf{25.45}& \textbf{24.13}\\
 \hline
\end{tabular}}
\label{tab:1}
\end{table}

\section {Conclusion}

This paper proposed a comparative study for the nuclear norms minimization methods.  We have devised an adaptive dictionary learning method to connect the rank minimization and GSR models. Based on the proposed adaptive dictionary, we have proved that NNM and WNNM are equivalent to $\ell_1$-norm minimization and the weighted $\ell_1$-norm minimization in GSR, respectively. Following this,  inspired by the correctness of enhancing sparsity of the weighted $\ell_1$-norm in comparison with $\ell_1$-norm in sparse representation, we have explained that WNNM is more effective than NNM. We have applied the GSR-WNNM model with image NSS prior to image denoising. Experimental results have demonstrated that WNNM is more effective than NNM and outperforms many state-of-the-art methods both quantitatively and qualitatively.

\section {Acknowledge}
This work was supported by the NSFC (61571102), the applied research programs of science and technology., Sichuan Province (No. 2018JY0035), the Ministry of Education, Republic of Singapore, under the Start-up Grant and the Macau Science and Technology Development Fund under Grant FDCT/077/2018/A2.

{\footnotesize
\bibliographystyle{IEEE}
\bibliography{gwnnm_ref}
}

\end{document}